\documentclass[journal]{IEEEtran}
\ifCLASSINFOpdf
\else
\fi
\pdfoutput=1 
\usepackage{graphicx}
\usepackage{lipsum}
\usepackage{caption}
\usepackage{subcaption}
\usepackage[super]{nth}
\usepackage{xcolor}
\usepackage{amsmath}
\usepackage{hyperref}
\usepackage{amssymb}
\usepackage[linesnumbered,ruled,vlined]{algorithm2e}
\begin{document}
%
\title{Reinforcement Learning Framework for Server Placement and Workload Allocation in Multi-Access Edge Computing}
%
%
%

\author{Anahita~Mazloomi,~Hani~Sami,~Jamal~Bentahar,~\IEEEmembership{Member,~IEEE,}~Hadi~Otrok,~\IEEEmembership{Senior~Member,~IEEE,}
        and~Azzam~Mourad,~\IEEEmembership{Senior~Member,~IEEE}
\thanks{Anahita~Mazloomi, Hani~Sami and Jamal~Bentahar (corresponding author) are with Concordia Institute for Information Systems Engineering, Concordia University, Montreal, QC, H3G 1M8, Canada (e-mail: ana\_mazloomi@yahoo.com; hani.sami@mail.concordia.ca; bentahar@ciise.concordia.ca).}
\thanks{Hadi~Otrok is with the Center of Cyber-Physical Systems, Department of EECS, Khalifa University, Abu Dhabi, UAE (email: Hadi.Otrok@ku.ac.ae).}
\thanks{Azzam Mourad is with the Department of Computer Science and Mathematics, Lebanese American University, Beirut 961, Lebanon (e-mail: azzam.mourad@lau.edu.lb).}}
\maketitle

\begin{abstract}
Cloud computing is a reliable solution to provide distributed computation power. However, real-time response is still challenging regarding the enormous amount of data generated by the IoT devices in 5G and 6G networks. Thus, multi-access edge computing (MEC), which consists of distributing the edge servers in the proximity of end-users to have low latency besides the higher processing power, is increasingly becoming a vital factor for the success of modern applications. This paper addresses the problem of minimizing both, the network delay, which is the main objective of MEC, and the number of edge servers to provide a MEC design with minimum cost. This MEC design consists of edge servers placement and base stations allocation, which makes it a joint combinatorial optimization problem (COP). Recently, reinforcement learning (RL) has shown promising results for COPs. However, modeling real-world problems using RL when the state and action spaces are large still needs investigation. We propose a novel RL framework with an efficient representation and modeling of the state space, action space and the penalty function in the design of the underlying Markov Decision Process (MDP) for solving our problem.
%
%
This modeling makesthe Temporal Difference learning (TD) applicable for a large-scale real-world problem while minimizing the cost of network design. We introduce the TD$(\lambda)$ with eligibility traces for Minimizing the Cost (TDMC) algorithm, in addition to Q-Learning for the same problem (QMC) when $\lambda=0$. Furthermore, we discuss the impact of state representation, action space, and penalty function on the convergence of each model.
%
%
%
Extensive experiments using a real-world dataset from Shanghai Telecommunication demonstrate that in the light of an efficient model, TDMC/QMC are able to find the actions that are the source of lower delayed penalty. The reported results show that our algorithm outperforms the other benchmarks by creating a trade-off among multiple objectives.
\end{abstract}


%
\IEEEpeerreviewmaketitle

\section{Introduction}
%
%
%
%
\IEEEPARstart{W}{ith} the development of IoT~\cite{Al-FuqahaGMAA15}, 5G and 6G networks, applications of smart mobile devices in different areas such as augmented reality, multiplayer games \cite{jia2015optimal}, image processing for facial recognition, natural language processing for real-time translation systems \cite{9039672} and data transferring among IoT devices on the internet of vehicles (IoV)~\cite{hammoud2020ai} have become more resource-intensive. More computation power and real-time response are needed, but mobile devices are limited in terms of the central processing unit (CPU), memory, storage, and processing power. 

First, mobile cloud computing (MCC) \cite{10.1145/2307849.2307856, DBLP:journals/monet/ZhangKJG11} was offered as a solution to overcome those constraints of handheld devices. Although cloud computing is a reliable solution to provide computation power, the real-time response is not guaranteed. The next solution was a cloudlet-based structure. Cloudlet is a group of powerful computers connected to the internet \cite{5280678}. The cloudlet has less computation power in comparison with the cloud, and because of the small coverage area, it is not scalable \cite{wang2019edge}.
Finally, mobile edge computing, currently known as multi-access edge computing (MEC), was offered as a novel network to mitigate the cloudlet shortcomings. It brings the computation power close to the end-users at the edge of the network, in a distributed manner \cite{9039672}. Thus, besides having more process power, the delay is decreased because of the computation source proximity. 

The mobile or IoT devices send a massive amount of requests to the edge servers, but there is a limited number of these servers because of the budget and energy consumption. Thus, the optimal locations for a limited number of edge servers among a large number of potential places should be found to have the minimum network delay. Besides, the computation power of the edge servers is not unlimited. Hence, discovering the dominant area of each edge server is required as each of these servers can only handle a certain number of requests. Overloading results in more delay compared to sending the requests to a farther server \cite{jia2015optimal}. Additionally, in assignments with idle edge servers, the network's cost increases, and the energy consumption of an idle edge server is about 60\% higher than a fully-loaded edge server \cite{8473378}. Therefore, strategic placement and optimal base station assignment are necessary to have a high quality of service (QoS) and quality of experience (QoE).

To solve the edge server placement problem, iterative/exhaustive search algorithms are space and time consuming. Due to the hardness of the problem and the possibility of having a large input size, the iterative search (such as K-means) is not guaranteed to produce efficient solutions in the case of multi-objective problems. Additionally, a heuristic-based solution is not guaranteed to reach an optimal solution even if the evolutionary-based algorithms are used.

Some studies have focused on task offloading by assuming that the edge servers' locations are known. In recent years, a number of them have considered the placement and task offloading together \cite{8975813}. This joint problem is an NP-hard combinatorial optimization problem (COP) consisting of searching and finding the optimal option among a limited set of discrete possible solutions. 
Recently, reinforcement learning (RL) \cite{sutton2018reinforcement} has shown promising results for COPs, including traveling salesman to find the shortest path, mixed-integer linear programming, the graph coloring problem, and the Knapsack problem  \cite{Mazyavkina2020ReinforcementLF}. However, there are a few studies that have considered the use of RL for placement optimization \cite{mirhoseini2020chip, mirhoseini2017device, murray2019adaptive, venkatakrishnan2019learning}. Although Q-learning is a powerful RL algorithm, in many real-world problems, finding the optimal solution is hard, especially when there is a large state or action space \cite{Mazyavkina2020ReinforcementLF, goldie2020placement,ChenLZSLWJHJL21}. In addition, Q-learning suffers from overestimation bias in some cases. Thus, we expand our solution to consider $n$-step learning using TD$(\lambda)$, where $\lambda$ is the trace decay parameter \cite{precup2001off}. If $\lambda=0$ (i.e., TD(0)), the solution becomes the classical Q-learning. In this case, our RL solution for MEC design that considers the Cost Minimization is called QMC. If $\lambda \in(0, 1]$, we call the solution TDMC.

Considering the placement and task offloading joint problem, most proposals have assumed that the number of edge servers is fixed and the main objective is minimizing the delay. 
In this paper, in addition to minimizing the delay, the main objective is to provide an efficient MEC design with minimum cost by minimizing the number of the edge servers. Minimizing the network delay along with the number of edge servers is a vital factor that should be considered towards having a competitive MEC design. This double objective should be attained under given constraints, including maximum accepted delay for the network and maximum workload capacity for edge servers. 
The output can be seen as different clusters of edge servers and their connected base stations. Providing an efficient RL framework for this capacitated clustering problem is our main focus. 
However, using RL for the server placement problem is highly challenging, and the number and distribution of the base stations, the delay and workload constraints, and the properties of edge servers are the factors that may increase the difficulty \cite{goldie2020placement}. Moreover, applying RL in real problems is not trivial; efficient modeling is needed to overcome the RL limitations for the problems with large and variable action space \cite{goldie2020placement, dulac2019challenges}. Furthermore, in our problem, there is no fixed situation as a goal for the agent, thus the reward value cannot be backtracked.

Hence, the contributions of this paper are as follows:
\begin{itemize}
    \item Providing a novel RL framework for the joint problem of edge server placement and workload allocation.
    \item Defining the state and action spaces, and penalty function for multi-objective problems without a fixed end-goal.
    \item Adapting the Temporal Difference learning with a backward-view mechanism (TD$(\lambda)$) for a real-world and scalable problem with a large number of base stations (2798). 
    \item Our solution outperforms a deep version of RL that needs more computational resources.
    \end{itemize}
    
   We conducted extensive experiments on the Shanghai Telecom dataset. The results show that our solution creates a reasonable trade-off between our two objectives and outperforms 
    other relevant benchmarks in the literature in terms of reducing the network delay and the number of edge servers. The benchmarks include the deep reinforcement learning using Deep Q-Network (DQN), Genetic Algorithm (GA), K-means, Top-K, K-means Top-K merged (KmTK), TopDoF, and Random.


In Section \ref{section1}, after presenting an overview of RL, Q-learning, and backward view TD, we review the most relevant related work. 
In Section \ref{section2}, the system model and our formulation are reported. Then in Section \ref{section3}, our RL framework for edge server placement and computation offloading is introduced and explained in detail. The issues and limitations of Q-learning and TD($\lambda$) implementations and the offered solutions are clarified. Section \ref{section4} shows extensive experiments, performance evaluations, and discussions after describing the real-world dataset. Moreover, the other methods which are used for performance comparison are explained. Finally, in Section \ref{section5}, a summary and some directions for future work conclude the paper.

\section{Background and Related Work} \label{section1}

\begin{figure}[ht]
    \centering
    \includegraphics[width=0.5\textwidth]{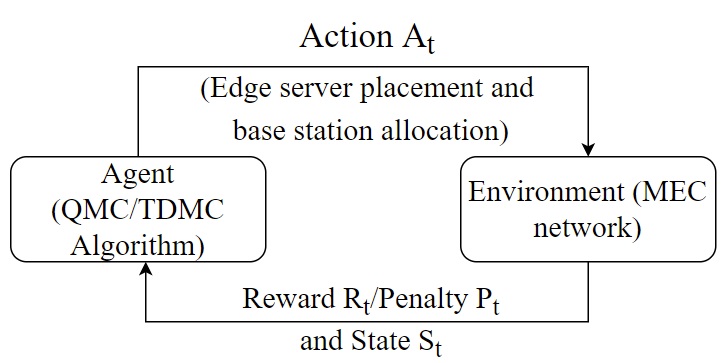}
    \caption{The reinforcement learning framework.}
    \label{fig:RL}
\end{figure}

RL is a type of machine learning with a learner called the agent. The agent makes decisions by interacting with its surrounding, i.e., the environment. The environment sends the reward or penalty signal and the new state to the agent after taking each action (see Fig.~\ref{fig:RL}). Through this continuous process, the agent learns to map situations to the actions. 

Generally, RL problems are formulated based on Markov Decision Processes (MDPs). Thus, in RL problems, state-space $\mathcal{S}$, action space $\mathcal{A}$, reward $\mathcal{R}$, state transition probability $\mathcal{P}$, and discount factor $\gamma$ should be defined through the MDP tuple $\langle \mathcal{S}, \mathcal{A}, \mathcal{R}, \gamma, \mathcal{P} \rangle$. 

Q-learning~\cite{watkins1992q} is one of the first major RL algorithms, which is a model-free, off-policy algorithm defined by the following Bellman equation:
\begin{equation}\label{eqn:Q}
\begin{split}
    & Q(S_{t}, A_{t}) \leftarrow  \\ & Q(S_{t}, A_{t}) + \alpha \times [R_{t+1} + \gamma \max_{a \in \mathcal{A}} Q(S_{t+1}, a) - Q(S_{t}, A_{t})]
\end{split}
\end{equation}

\noindent where $S_{t}, S_{t+1} \in \mathcal{S}$ are states at time $t$ and  $t+1$, $A_{t} \in \mathcal{A}$ is the action at time $t$, and $R_{t+1} \in \mathcal{R}$ is the immediate reward after taking each action. The Q-values represent the quality of actions in each state, and $\alpha$ is the learning rate that has a value between 0 and 1. These Q-values are stored in a tabular (Q-table) format where the rows and columns consist of the states and actions respectively. 
Classic tabular learning updates involve using TD(0) (Q-learning) or TD(1) which is a Monte-Carlo implementation \cite{precup2001off}. Despite the good performance entailed by each method, they suffer from bias and variance issues towards the action-value function (Q-value) update. To overcome this problem, we utilize TD$(\lambda)$ for solving our problem. TD$(0)$ involves a one-step Q-value update, while TD$(1)$ waits until the end of the episode to update the Q-value of the traversed trajectory. A better solution is to consider an intermediate number of steps, which is specified by the \textit{trace decay parameter} $\lambda \in [0, 1]$ and using the backward-view of TD$(\lambda)$. In order to implement the backward-view, eligibility traces $e_t(s,a)\in\mathbb{R}^+$ is introduced for each state $s$ and action $a$ in a separate table. At each step, the eligibility trace for the visited state is increased by $1$, while it decays for all the states by $\gamma^{\lambda}$ as follows:
\begin{equation}
e_{t+1}(s, a) = 
 \begin{cases}
    \gamma \lambda e_{t}(s, a), & \text{if }A_t=a, S_t=s\\
    \gamma \lambda e_{t}(s, a) + 1, & \text{ otherwise}
 \end{cases}
\end{equation}
Using traces, we are able to tell the degree of which a state is eligible to undergo learning changes based on the reinforcing event. The TD error performs proportional learning update to the visited states. Therefore, Equation \ref{eqn:Q} is transformed into:
\begin{equation}
     Q(S_{t}, A_{t}) \leftarrow Q(S_{t}, A_{t}) + \alpha \times \mathcal{E}_t e_t(S_t, A_t)
\end{equation}
where $\mathcal{E}$ is the error term at time $t$. 
In this approach $\lambda$ controls the trade-off between bias and variance, and is guaranteed to influence the learning speed and the asymptotic performance in the function approximation.

By assuming the known locations for the servers in~\cite{dab2019q}, Q-learning is used to find the optimal offloading policy where the objective is to reduce energy consumption with a fixed maximum delay as a constraint. After modeling the offloading as an MDP, the reward is calculated based on the time and energy consumption that should be minimized. 
Sen and Shen~\cite{sen2019machine} also used Q-learning with three actions in each state. The arrival task can be processed on the edge, fog, or cloud. After the convergence of the Q-value, the Q-table is used for task allocation in edge servers. The authors showed that their approach could reduce execution time and energy consumption.
For the joint optimization of task offloading and bandwidth allocation, in~\cite{huang2019deep,zhao2020deep}, a deep-Q network is used to learn the offloading policy where the objective is minimizing both the latency and energy consumption.

To define the optimal number of cloudlets to have a trade-off between the QoS and the cost for the service provider, Peng~\textit{et al.}~\cite{9001043} used an improved affinity propagation algorithm. In their clustering, they considered user movement and load balancing. They divided the density of mobile users before and after moving into three clusters: sparse, discrete, and dense. The place of cloudlets would be changed by the density to cover more users.
The authors generated a dataset for different numbers of mobile users, and their algorithm outperformed K-means and mini-batch K-means.
Wang~\textit{et al.}~\cite{wang2019edge} have proposed the first study for the edge server placement. They used mixed integer programming (MIP) for edge servers placement and base station allocation by considering the access delay and load balancing as the two objectives for their formulation. It is considered that the fixed number of identical edge servers is given. Their results, on Shanghai Telecom's dataset, are compared with the K-means, top-K, and random approaches for different numbers of base stations on a large scale, for 300 to 3000 base stations. Although their experiments showed K-means has less delay and top-K creates a more balanced cluster, in total, their approach has better results. In~\cite{guo2020user}, for the same dataset, they have used the MIP and K-means combination for 20 to 200 base stations. Authors in~\cite{XuXQZWDC19} used a three-step approach to find the edge server placement to minimize the delay while considering the workload balancing. They used a decision tree in the first level with the help of a genetic algorithm and multiple criteria decision-making techniques in the following steps. They compared their proposed approach with the two other greedy methods, where the nearest edge server or the edge server with the most available computing power is selected.
To find the optimal placement and resource allocation in MEC, various approaches are investigated, including mainly: clustering~\cite{9001043}, top-K~\cite{jia2015optimal}, MIP~\cite{wang2019edge}, combination of MIP and K-means~\cite{guo2020user}, heuristics~\cite{li2018energy, XuXQZWDC19}, ILP and game theory~\cite{article}. 

There are a few studies that have considered RL in placement optimization.
Mirhoseini~\textit{et al.}~\cite{mirhoseini2017device} proposed a deep RL algorithm to find the optimal placement for different operations of neural networks onto hardware devices that can be CPU or GPU. They used the square root of the execution time as a reward, and they demonstrated a reduction in the single-step running time as well as the total training time compared to the heuristics and traditional methods. Addanki~\textit{et al.}~\cite{venkatakrishnan2019learning} studied the same problem using the graph embedding method that helped generalize and improve the algorithm to be applicable for different neural networks, rather than only a specific one. The algorithm's output is a policy instead of places for operations. This learned policy is transferable for the unseen neural networks of the same family. 
In both~\cite{mirhoseini2017device, venkatakrishnan2019learning}, before implementing the RL, the authors have grouped the same operations and forced the algorithm to place them in the same device, which reduces the placement actions. Mirhoseini~\textit{et al.}~\cite{mirhoseini2020chip} used RL for chip placement. The performance of their algorithm improves over time by having more experience. As in~\cite{venkatakrishnan2019learning}, their algorithm can be employed for unseen blocks, even when these blocks have bigger size. However, unlike~\cite{venkatakrishnan2019learning}, in~\cite{mirhoseini2020chip}, the placement policy is defined incrementally after each state until reaching the terminal. The reward for each step is zero, except for the final one. To consider all the objectives, the authors used a weighted sum in a single reward function, which is computed by a neural network. This neural network is made by a rich state representation and a large dataset to train it in a supervised manner. They have noticeably improved the placement time while outperforming other methods.

Applying RL for edge server placement still needs investigation. There is no fixed point to achieve as a goal and minimizing the path's penalty is the objective. Our agent needs to find the optimal nodes as edge servers that are fixed and do not vary over time. Therefore, in this paper, a novel TD$(\lambda)$-based approach with a backward-view mechanism is examined and applied to design MEC with regard to optimized placement of edge servers while considering the deployment and delay cost as variables to be minimized. 

\section{Edge Server Placement Modelling}\label{section2}

\subsection{System Model and Problem Definition}

MEC can be shown by a set of $n$ base stations $BS =$ \{$bs_{1}, bs_{2}, \dots, bs_{n}$\} and a set of $K$ edge servers $ES =$ \{$es_{1}, es_{2}, \dots, es_{K}$\}, ($K$ $<$ $n$). Edge server placement in 5G and 6G networks can be shown by an undirected graph $G = (V, E)$, where $V = BS\bigcup ES$ and $E$ is the representation of the connections between base stations and edge servers. Usually, for  simplicity purposes, it is assumed the edge servers are co-located with the existing base stations \cite{wang2019edge,9001043}.
\begin{figure}
    \centering
    \includegraphics[width=0.5\textwidth]{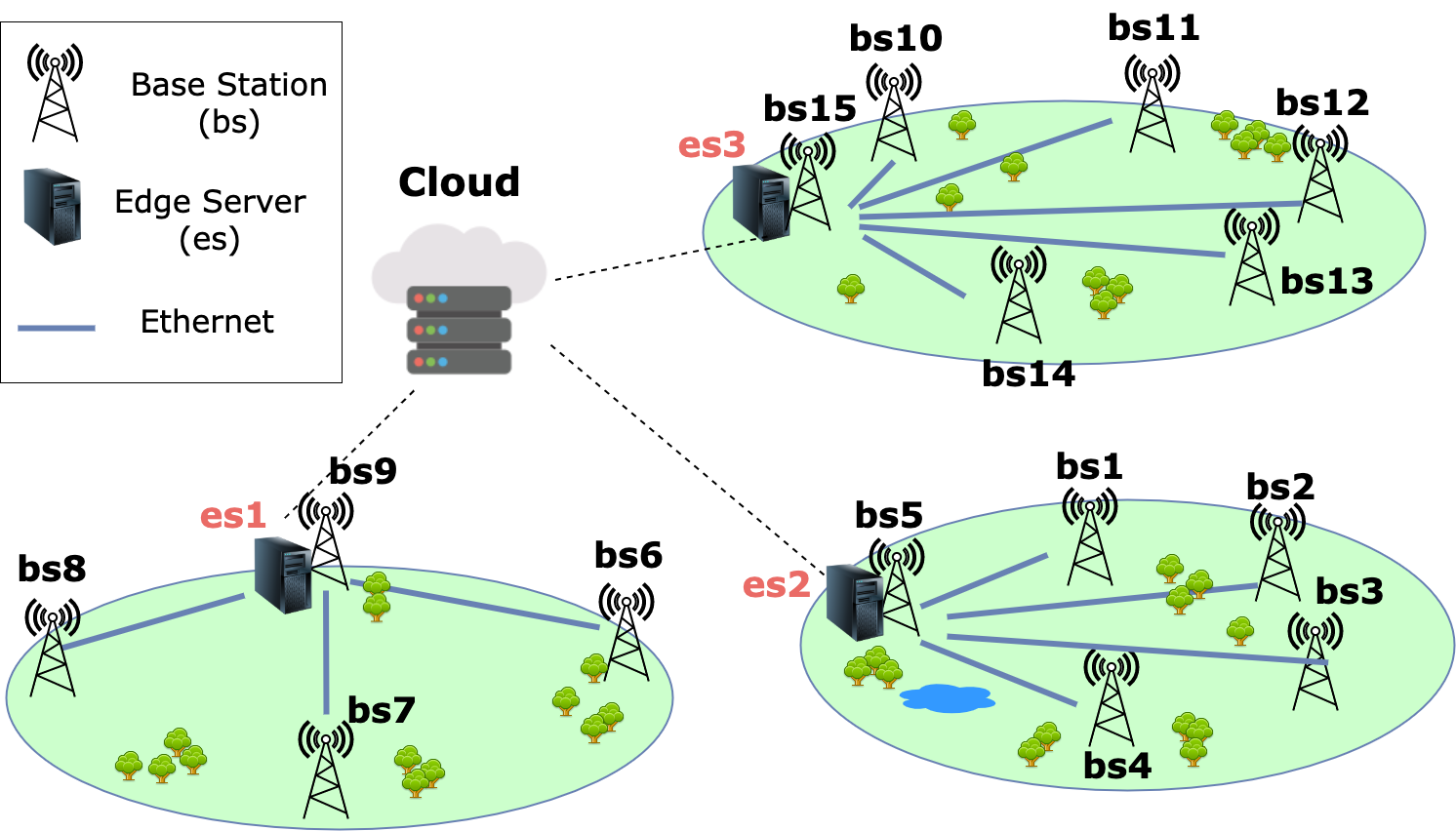}
    \caption{Edge-server placement in MEC.}
    \label{fig:MEC}
\end{figure}

In Fig.~\ref{fig:MEC}, the bigger base stations illustrate the equipped base stations with servers. For example, $es1$ is the edge server that covers base stations $bs6$, $bs7$, $bs8$, and the co-located base station $bs9$. The workload of $es1$ is the total workload of all connected base stations. 
In order to have an optimal network design, the following assumptions are considered in our model:
\begin{itemize}
    \item Each edge server is co-located with one of the existing base stations.
    \item Each edge server covers many base stations while each base station is only connected to one edge server, and the selected edge servers should cover all the base stations.
    \item The distance between the base station and its edge server reflects the delay.
\end{itemize}


The MEC design refers to finding the optimal locations to add the servers, and base station allocation/workload offloading by using the historical data from users' connections to the base stations over days. These two problems are known to be NP-hard \cite{wang2019edge}, therefore we are proposing an RL-based algorithm to solve them. The key variables used in our formulations and modeling are shown in Table \ref{table:Notations}. 

\begin{table}
  \begin{center}
    \caption{Notations.}
    \label{table:Notations}
    \begin{tabular}{|p{0.1\linewidth}|p{0.8\linewidth}|}
    \hline
     Symbol & Meaning \\
    \hline
     $BS$ & set of all the base stations in the network \\ 
     $bs_{i}$ & base station $i$ in the network, 1$\leq i \leq$ $n$ \\
     $ES$ & set of all the edge servers in the network \\
     $es_{j}$ & edge server $j$ in the network, 1$\leq j \leq$ $K$ \\
     $G$ & mobile edge computing network \\
     $K$ & number of edge servers\\
     $D_{N}$ & Network delay\\
     $n$ & number of base stations\\
     $|n_{j}|$ & number of base stations in cluster $j$\\
     $\overline{D}$ & set of average distances of all clusters\\
     $\overline{D}_{j}$ & average distances of base stations, in cluster $j$, from the $j^{\text{th}}$ edge server\\
     $D_{TH}$ & distance threshold\\
     $d_{ij}$ & distance of base station $i$ from edge server $j$\\
     $\Delta$ & set of all edge servers' workloads in the network \\
     $\Delta_{j}$ & workload of edge server $j$ in the network \\ 
     $\delta$ & set of all base stations' workloads in the network \\
     $\delta_{i}$ & workload of base station $i$ in the network \\
     $x_{ij}$ & binary variable, $x_{ij}=1$ if base station $i$ is connected to the edge server $j$\\ 
     $L_{bs_{i}}$ & location of the base station $i$\\
     $lat_{bs_{i}}$ & latitude of a base station $i$\\
     $lon_{bs_{i}}$ & longitude of a base station $i$\\
     $a_{bs_{i}}$ & action space of $bs_{i}$\\
     $\alpha$ & learning rate\\
     $\gamma$ & discount factor\\
     $ne_{1_{bs_{i}}}$ & $1^{\text{th}}$ neighbor of $bs_{i}$ with respect to the distance\\
     $wl_{ES}$ & dictionary of the edge servers' workloads \\
     $wl_{j}$ & workload of each edge server $j$, 1$\leq j \leq$ $K$\\
     $z$ & binary variable if selected base station is a new edge server $z = 1$\\
     $Pr_{i}$ & priority of $bs_{i}$\\
    \hline
    \end{tabular}
    
    \end{center}
\end{table}

\subsection{Computational Problem} 

Our objective is to minimize the network's cost, which relies on two parts: minimizing the number of edge servers $K$ (Eq.~\ref{eqn:O1}) and minimizing the network delay $D_{N}$ (Eq.~\ref{eqn:O2}). Deployment of new edge servers increases the cost of the network, and as mentioned having a minimum delay is one of the most vital objectives that should be considered. In fact, as argued in~\cite{9039672}, the long execution time is one of the main concerns in the design of MEC. Therefore, this problem is modeled as follows: 
\begin{equation}\label{eqn:O1}
     O_{1}: \min (K)
\end{equation}
\begin{equation}\label{eqn:O2}
    O_{2}: \min (D_{N})
\end{equation}
subject to the following constraints:

\begin{equation}\label{eqn:Con11}
    \Delta_{j} \leq \Delta_{max}\hspace{10pt}(\forall j , 1 \leq j\leq K)
\end{equation}
\begin{equation}\label{eqn:DNC}
    \sum_{j=1}^{K}(x_{ij}d_{ij}) \leq D_{TH} \hspace{10pt}(\forall i , 1 \leq i\leq n)
\end{equation}
\begin{equation}\label{constraint1}
 \displaystyle \sum_{j=1}^{K}x_{ij} = 1\hspace{10pt}(\forall i , 1 \leq i\leq n)
\end{equation}
\begin{equation}\label{constraint111}
    x_{ij}\in \{0, 1\}\hspace{10pt}(\forall i , 1 \leq i\leq n\hspace{5pt} and \hspace{5pt} \forall j , 1 \leq j\leq K)
\end{equation}
The first constraint (Eq.~\ref{eqn:Con11}) states each edge server has a limited computational power. An upper bound, $\Delta_{max}$, is considered for the edge servers as the processing capability. As the distance reflects the delay, $D_{TH}$ is used to represent the distance threshold representing the maximum acceptable delay (Eq.~\ref{eqn:DNC}). It is worth noting that, although an acceptable constraint for the latency is considered (Eq.~\ref{eqn:DNC}), as mentioned above in Eq.~\ref{eqn:O2}, still our goal is always to further minimize the delay. Knowing that $D_{N_i,j} = \frac{d_{i,j}}{Speed_{i,j}}$, the delay is derived from the distance, where $D_{N_i,j}$ and $Speed_{i,j}$ are the networking delay and propagation speed between $i$ and $j$ respectively.

Finally, each base station is connected to just one edge server while all the base stations are covered (Eq.~\ref{constraint1}), $x_{ij}$ is a binary variable (Eq.~\ref{constraint111}) and $x_{ij} = 1$ if base station $i$ is connected to the edge server $j$; otherwise, $x_{ij} = 0$.

\section{Algorithmic Implementation} \label{section3}

The joint problem of finding the optimal placement and workload offloading is modeled as an RL problem. Moving forward from one base station to the other one, from 1 to $n$, is the path in each episode. At each time step, the agent is in the location of one of the base stations. Then, the agent takes action for the base station and receives the result in the form of a penalty. This procedure continues until the agent moves through all the base stations. After enough iterations and experiencing different interactions with the environment, the agent could be able to find the actions with the minimum penalty for the path.

\subsection{Initial State Space}
As RL is based on MDP, defining the state space, action space, and the reward function are the fundamental steps. First, the state space is defined and the agent moves forward from one base station to the next one. Thus, the state-space is a set of locations of all base stations because the agent locates in the place of one of them at each time step. This location has two geographical coordination: latitude and longitude. The initial state and the location are formalized as follows:

$Initial\_State\_set = [\ L_{bs_{1}},\ L_{bs_{2}}, \dots,\ L_{bs_{n}}\ ]$

 $L_{bs_{i}} = [\ lat_{bs_{i}},\ lon_{bs_{i}}\ ]$
\subsection{Action Space}
Second, by considering the predefined distance threshold based on the delay budget, the action space is defined. We create an adjacency matrix of zeros and ones for our network. It is a square matrix where rows and columns show the base stations. In this symmetric matrix, the entries are zero unless their value in the distance matrix is less than the threshold. The distance matrix has the same dimension, but the values show the distance between the adjacent base stations. If there are $n$ base stations, this matrix will be $n\times n$. Then the Hadamard product of each row of the adjacency matrix and the joint action is the neighbors vector for each of the base stations. The neighbors vector shows the possible action space for each base station. Finally the non-zero elements are sorted based on the distance matrix and the result is the base station’s action space. Therefore, the action space for each base station is limited to the nodes with value one in the adjacency matrix (i.e., the neighbors). The joint action space of all base stations is:

$A = \{a_{bs_{1}},a_{bs_{2}}, \dots , a_{bs_{n}} \} $

$a_{bs_{i}} = [ne_{1_{bs_{i}}},ne_{2_{bs_{i}}}, \dots,ne_{N_{bs_{i}}}]; \hspace{5pt}$

\noindent where $a_{bs_{i}}$ shows the action space of base station $i$ which is limited to its neighbors and $ne_{1_{bs_{i}}}$ is the first neighbor of base station $i$. The size of this action space varies for each node, and it is equal to the number of the neighbors (N) of each base station. For instance, assuming there are five base stations and the distance matrix is as follows:
\begin{align*}
    \begin{bmatrix}
        0&5&8&9&4\\
        5&0&3&17&2\\
        8&3&0&10&14\\
        9&17&10&0&20\\
        4&2&14&20&0\\
    \end{bmatrix}
\end{align*}
In case $D_{th} = 7$, the adjacency matrix becomes:
\begin{align*}
    \begin{bmatrix}
        1&1&0&0&1\\
        1&1&1&0&1\\
        0&1&1&0&0\\
        0&0&0&1&0\\
        1&1&0&0&1\\
    \end{bmatrix}
\end{align*}
Afterwards, the neighbors vector for each base station is computed based on the Hadamard product of the first row in adjacency matrix and the joint action space:
\begin{align*}
    &\begin{bmatrix}
        1&1&0&0&1\\
    \end{bmatrix}
    \circ
    \begin{bmatrix}
        bs_1&bs_2&bs_3&bs_4&bs_5\\
    \end{bmatrix}
    =\\
    &\begin{bmatrix}
        bs_1&bs_2&0&0&bs_5\\
    \end{bmatrix}
\end{align*}
Finally, the set of feasible actions for $bs_1$ becomes:
\begin{align*}
    a_{bs_1} =
    \begin{bmatrix}
        bs_1, bs_5, bs_2
    \end{bmatrix}
\end{align*}

Variable action space is a challenging situation in deep reinforcement learning \cite{boutilier2018planning, chandak2020reinforcement}. This issue arises when function approximation should be used. Hence, It is preferred to model the problem in a way that can be solved by Q-table.

\subsection{Initial Penalty Function}

Next, the environmental signal, which is sent to the agent as the penalty ($P$), is determined. This signal leads the agent toward taking the optimal actions in the given states. The objective is minimizing the cost, which consists of delay and the number of base stations.

We align the standard Q-learning formula to our objective function as follows:
\begin{equation}\label{eqn:QP}
\begin{split}
    & Q(S_{t}, A_{t}) \leftarrow \\ & Q(S_{t}, A_{t}) + \alpha \times [P_{t+1} + \gamma \min_{a \in \mathcal{A}} Q(S_{t+1}, a) - Q(S_{t}, A_{t})]
\end{split}
\end{equation}

\noindent $P_{t+1}$ represents the immediate penalty, and the agent's policy is to minimize the penalty, instead of maximizing the reward. We considered $\alpha = 0.4$ and $\gamma = 0.9$ for our algorithm.

Goldie and Mirhoseini~\cite{goldie2020placement} assert that penalty/reward calculation for the placement problems when the action space is not completed (partial placement) is very difficult. Further, for multi-objective problems, it is usually hard to determine a single penalty value. And to evaluate the policy, it is crucial to consider the performance of the algorithm for each objective separately~\cite{dulac2019challenges}.

\subsection{Challenges and Solutions}
Distance, 
as the first factor, was used in calculating the penalty because it reflects the delay in our modeling. By having this penalty function, after iterations, the action in each base station (state) is choosing itself as the destination. In this situation, the distance is zero, which minimizes the penalty. This result is in contrast with the other objective (Eq.~\ref{eqn:O1}). Therefore, a balance between the two goals is needed.

To address the above issue, a fixed value, after having a new edge server, is added to the penalty. Setting this value is challenging and should be defined based on the model's constraints. In our algorithm, it should be higher than the distance threshold. Otherwise, the agent would add extra edge servers to the network while it could select one of the existing ones in the accepted distance. For example, when our distance threshold was 9~\textit{km}, this fixed value was considered 10. This motivated the agent to select existing edge servers within 9~\textit{km} of the current base station rather than adding a new edge server because of receiving less penalty.  

Although adding this value helps the agent take better actions, it is not enough for our algorithm to converge to the minimum cost. It is because of the actions' dependency and the lack of a static final goal. For example, to find the optimal route in the maze problem\footnote{\url{https://www.samyzaf.com/ML/rl/qmaze.html}}, there is a point when reached, it brings a high reward; otherwise, the length of the path is used as a reward. Then, the agent can learn the optimal policy by backtracking the reward value. However, in our allocation problem, there is no end point with higher reward and the the length of the path is the same in all episodes because the workload offloading should be done for all the base stations. When the agent reaches the base station $n$, the episode finishes, but it is not the final goal. Therefore, the agent is not able to find a systematic rule to select the optimal action based on Q-values changes and thus, convergence is not guaranteed.

Moreover, as the length of the path increases, the agent faces the credit assignment problem (CAP)~\cite{minsky1961steps}. The agent is not able to define which action is resulting in a higher penalty because, in our problem, the lower penalty in the current state does not guarantee the minimum cost (penalty) for the completed path or even in the next episode.  

\subsubsection{Modified State Space}
To address the two aforementioned issues: lack of end goal and CAP, a list of previously taken actions in each episode is added to the state definition at each time step. In other words, the list of edge servers up to the current time step is appended to the state space as follows:

$Modified\_State_{t} = [ L_{bs_{t}}, [a_{s_{1}}, a_{s_{2}}, \dots, a_{s_{t-1}} ]]$

For completing the penalty function after a precise definition of state and action space, constraints should be considered. Penalizing the actions that do not satisfy the constraints with a high penalty is a possible solution. In this case, the agent could be trapped, if there are numerous infeasible actions. Additionally, there might be a probability of not satisfying the constraints, and it needs more time and more iterations to reach the optimal value. The other method that can be applied is to prevent the agent from taking forbidden actions by setting some rules. In our study, the latter approach is chosen.   
\subsubsection{State Space}
For the distance constraint, as explained previously, the action space is limited to the neighbors. For the workload constraints, a dictionary of the current edge servers and their workloads is added to the state definition ($wl_{ES}$), which gets updated after each action. Therefore, our complete state definition at time step $t$ is:

$Final\_Modified\_State_{t} = [ L_{bs_{t}},\ [a_{s_{1}},\ a_{s_{2}},\ \dots,\ a_{s_{t-1}}],\\ \ \{es_{1} : wl_{1}, es_{2} : wl_{2}, \dots \}]$

This definition simplifies the penalty calculation, shrinks the action space further, and keeps the agent away from taking infeasible actions. 

\subsubsection{QTable}
There is another situation that the agent should avoid. For instance, if the agent in $bs_{5}$ selects $bs_{10}$ as the destination for offloading, it cannot map the other base stations to $bs_{5}$. For this reason, a list of forbidden actions ($FA$) is created. As the agent moves forward, the list expands, and the action space becomes smaller, but it causes variable action space for each state in each episode, which, as mentioned earlier, is a highly challenging problem for a deep Q-network (DQN). Besides, the following reasons  make DQN unsuitable for our problem: (1) each input with a different size might require a new network design; (2) DQN requires a large number of hyperparameters to be tuned for each configuration such as mini-batch size, replay buffer size, neural network configuration, optimizer configuration, loss function configuration, etc.; and (3) our problem does not require a non-linear approximation because the states do not change over time. As shown in the evaluation section (Section \ref{DRL-DQN}), the DQN solution is not capable of converging to the minimum cost compared to our proposed TD$(\lambda)$ solution. 

Our modeling solves the aforementioned difficulties, namely: large and variable action space, penalty calculation, CAP, and lack of exact end goal. However, to use the Q-table, the large state space challenge has to be addressed. In the Q-table, the states are the rows, and the actions are columns. In our state definition, there is a fixed part, the base station location, which is the same in all the episodes. The other components vary based on different actions. In our Q-table, as the final goal is to have the best action for each base station, only the fixed part is considered as the state to be fitted in the look-up table (Algorithm~\ref{alg:QMC-alg}. line 6). It should be noted that in different episodes, the agent uses the variable parts, the selected edge servers and their workloads, to calculate the penalty and take the appropriate actions. Thus, the variable parts function as a memory for our agent in the decision-making process because the fixed section is not able to represent the whole information of each state. Therefore, the $n \times n$ matrix is created as our Q-table. Initially, it is filled out with a high value of $1000$. Then, the entries with value one in the adjacency matrix change to zero (Algorithm~\ref{alg:QMC-alg}. line 1). These initial action values motivate the agent for exploration, and the possible actions would be tried several times before the convergence \cite{sutton2018reinforcement}.
%
%

\subsubsection{Penalty Function}
Having completed the model, the remaining issue is that the actions are dependent on the node orders. For example, the agent always selects the first node as an edge server. If the order of nodes changes, the edge servers will change as well. To tackle this problem, a priority value, $Pr$, is created for each node. It can partially show the importance of nodes for being selected as the edge server. For example, selecting a node with a higher workload is preferred because it reduces the workload transferring, which results in less delay. Also, if a node is in a dense area, it is a more suitable place for adding the edge servers. The priority $Pr_i$ of the base station $i$ is expressed as follows: 
\begin{equation}\label{eqn:Priority}
    Pr_{i}= (\delta_{i} + DoF_{i})/{average\_distance}_{i}; \hspace{5pt} (\forall i, 1 \leq i \leq n)
\end{equation}
\noindent where $\delta_i$ is the workload of base station $i$. $DoF_{i}$ is the degree of freedom that shows the number of neighbors situated in the accepted distance. $DoF$ represents the directions that each base station can transfer its computation or receive tasks from other nodes. ${average\_distance}_{i}$ is the average distance of $N(=15)$ nearest nodes from the base station $i$. The inverse of the $Pr$ expression is the penalty of selecting each node, which helps the agent make decisions regarding the importance of the nodes not based on their orders.

Therefore, the penalty function is:
\begin{equation}\label{eqn:Penaltyyy}
    P = distance (L_{bs_{i}}, L_{a_{bs_{i}}}) + fixed\_value \times z + 1/Pr_{a_{bs_{i}}}
\end{equation}

\noindent where the distance is calculated based on Eqs.~\ref{eqn:dlon}-\ref{eqn:calculatedistance}, which will be explained in Section~\ref{section4}.
The input of the distance function, $distance (L_{bs_{i}}, L_{a_{bs_{i}}})$, is the location of the current node, $L_{bs_{i}}$, and the location of the selected node for computation offloading. The action of the current state is also a base station, and its location ($L_{a_{bs_{i}}}$) should be used in Eq.~\ref{eqn:Penaltyyy}. The fixed\_value is added if the selected node has not been in the list of the edge servers ($ES$) which is the reason for considering the $z$ as a binary variable. The last part is the inverse of the priority of the selected destination.

\subsection{Initialization and Exploration}
So far, the fundamental parts for creating the efficient TD learning algorithm are explained. An additional step is needed to make this efficient modeling applicable to our problem. In each episode, a list of possible actions ($PA$) for each state is created. Initially, the list is empty, and as the agent passes the states, it fills based on the actions. For example, in each state, this list is a subset of the current edge servers that are within the current state's acceptable distance. Then, based on the $\epsilon$-greedy policy, the agent in each state selects the actions from the possible-action list or the action space that was defined for the classic Q-learning algorithm. The parameter $\epsilon$, which represents the probability of selecting a random action, is calculated as follows:
\begin{equation}\label{eqn: ep}
   \epsilon = 9/(T +100)
\end{equation}
\noindent where $T$ is the number of episodes. In the first iterations, the action selection is rather based on exploration. After some iterations, the agent relies on exploiting the best-offered actions based on the Q-value. Consequently, the possible-action list, based on this policy, is improved as well.

\subsection{TDMC Algorithm}
Algorithm~\ref{alg:QMC-alg} represents our proposed TD$(\lambda)$ framework for minimizing the cost (TDMC), and as mentioned earlier, when $\lambda$ = 0, the algorithm becomes QMC (Q-learning for minimizing the cost). \textsc{ResetEnvironment()} refers to the process of updating the edge servers list in each episode ($ES$ = [ ]), updating the workload dictionary of the selected edge servers ($wl_{ES}$ = \{ \}) and updating the forbidden action's list ($FA$ = [ ]). Lines 6 and 7 force the agent to follow the mentioned rule that a node cannot be both sender and receiver.

\vspace{1em}
\newcommand\mycommfont[1]{\footnotesize\ttfamily\textcolor{blue}{#1}}
\SetCommentSty{mycommfont}

\SetKwInput{KwInput}{Input}                
\SetKwInput{KwOutput}{Output}              

  
    


\begin{algorithm}
\DontPrintSemicolon
  
  Initialize $Q(s, a)$ 
  to $1000$ except adjacent nodes for each state to 0, and $Q(terminal, .) = 0$ \\
  Initialize $e(s, a)$ to $0$\\
  \For{episode $\leftarrow$ 1 to MAX}    
        { 
        	\textsc{ResetEnvironment()}\\
        	\For{i $\leftarrow$ 1 to $n$}    
        { 
        	$s_{i}$ $\leftarrow$ $L_{bs_{i}}$\\
        	
        	  \If{$s_{i}$ in ES}
    {
        $a_{bs_{i}}$ $\leftarrow$ $bs_{i}$, obtain $P$, $s_{i+1}$ $\leftarrow$ $L_{bs_{i+1}}$, update $Q$, update $e$, update $ES$, update $wl_{ES}$\\   
        
    }
    \Else
    {
        $PA$ $\leftarrow$ \textsc{GetPossibleActions()}\\
    $Q\_PA$ $\leftarrow$ \textsc{GetQvaluesofPA()}\\
    \If{$|PA|$ $>$ 0}
    {
    	Choose $a_{bs_{i}}$ for $s_{i}$ using $\epsilon$-greedy policy from $Q\_PA$\\
    }

    \Else
    {
        $Nei$ $\leftarrow$ \textsc{GetNeighbors()}\\
    $Q\_Nei$ $\leftarrow$ \textsc{GetQvaluesofNei()}\\
    	Choose $a_{bs_{i}}$ for $s_{i}$ using $\epsilon$-greedy policy from $Q\_Nei$\\
    }
    take action $a_{bs_{i}}$, obtain $P$, $s_{i+1}$ $\leftarrow$ $L_{bs_{i+1}}$, update $Q$, update $e$, \textsc{UpdateEnvironment()} \\
    	\tcc{UpdateEnvironment(): update $ES$, update $wl_{ES}$ and update $FA$}}
    }

        }

\caption{TDMC Framework}
\label{alg:QMC-alg}
\end{algorithm}

\section{Results and Discussion} \label{section4}
To evaluate the performance of our proposed algorithms, a dataset from Shanghai Telecommunication~\cite{wang2019edge, wang2019qos, guo2020user, wang2019delay} in China is used. It comprises the information related to 3233 base stations and the connected users over 30 days, in June of 2014. The number of requests is used for workload calculation. The number of all the requests from different users that are directed to each base station every day represents the base station's workload. After computing this value for each of the base stations over 30 days, the maximum value is the base station's workload, $\delta =$ \{$\delta_{1}, \delta_{2}, \dots, \delta_{n}$\}. Thus, in the workload calculation, the worst situation is considered because the maximum workloads may not happen concurrently. The reason is that the placement of selected nodes as edge servers does not change over a short period of time. Moreover, because of the fast growth of the number of mobile and IoT devices, this conservative workload calculation is preferred. Then, based on the capacity limitation of the edge servers, the maximum workload that each edge server can handle was considered 150 requests per day similarly to the setup in~\cite{8939566}. Considering the base stations' workload distribution
, some of them with higher workloads were omitted from our dataset.

As mentioned, distance represents the delay. The distance matrix is created using the geographical locations in the dataset. For that, the fourth column of the dataset is divided into two separate columns. Instead of Euclidean distance that is the length of a straight line between two points, the following equations are used to have more precise distances:

\begin{equation}\label{eqn:dlon}
dlon = lon2 - lon1 
\end{equation}
\begin{equation}
\label{eqn:dlat}
 dlat = lat2 - lat1
\end{equation}
\begin{equation}
\label{eqn:haversine}
a = (sin(dlat/2))^2 + cos(lat1) \times cos(lat2) \times (sin(dlon/2))^2
\end{equation}
\begin{equation}
\label{eqn:greatcircle}
c = 2 \times atan2( sqrt(a), sqrt(1-a))
\end{equation}
\begin{equation}
\label{eqn:calculatedistance}
d = R \times c
\end{equation}

\noindent where $lon$ and $lat$ represent the longitude and latitude of each node in radians, respectively. Eq.~\ref{eqn:haversine} is the haversine formula that computes the great-circle distance between two points in Eq.~\ref{eqn:greatcircle}~\cite{sinnott1984virtues}. This distance is the shortest path between the two points on the surface of the sphere. The $atan2(x, y)$ is the arctangent that gives the radians angle between $x$ and $y$. And $R$ is the radius of the earth that is equal to 6371~\textit{km}. In the last step, the base stations which have fewer than five neighbors in their 3~\textit{km} distance are known as the outliers and are removed from the dataset.

In our constrained MDP, two constraints of maximum workload and maximum distance need to be defined before implementation. The maximum capacity of each edge server is considered equal to 150 requests per day, and the maximum acceptable delay for our network is 0.03~\textit{ms}. To satisfy the delay constraint, the distance between the edge server and the dominated base stations should be within 9~\textit{km}~\cite{8939566}.
\begin{figure*}\label{fig:100basestations1}
    \centering
    \begin{subfigure}{0.18\textwidth}
        \includegraphics[width=\textwidth]{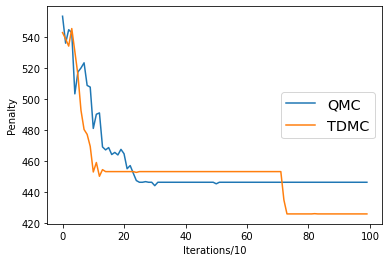}
        \caption{100 base stations}
        \label{fig:100basestations}
    \end{subfigure}
    \hfill
    \begin{subfigure}{0.18\textwidth}
        \includegraphics[width=\textwidth]{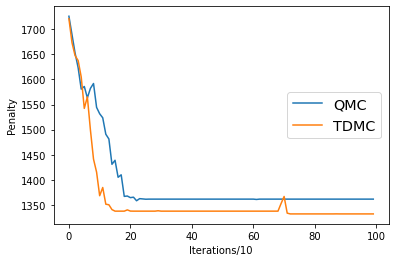}
        \caption{300 base stations}
        \label{fig:300basestations}
    \end{subfigure}
    \hfill
    \begin{subfigure}{0.18\textwidth}
        \includegraphics[width=\textwidth]{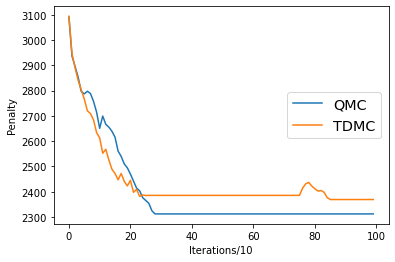}
        \caption{500 base stations}
        \label{fig:500basestations}
    \end{subfigure}
    \hfill
    \begin{subfigure}{0.18\textwidth}
        \includegraphics[width=\textwidth]{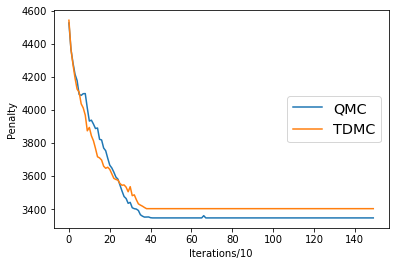}
        \caption{700 base stations}
        \label{fig:700basestations}
    \end{subfigure}
    \hfill
    \begin{subfigure}{0.18\textwidth}
        \includegraphics[width=\textwidth]{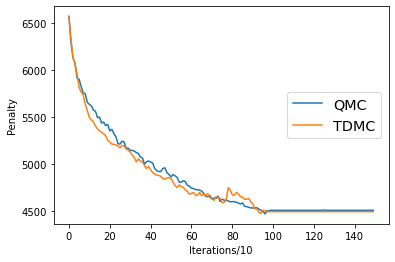}
        \caption{1000 base stations}
        \label{fig:1000basestations}
    \end{subfigure}
    \begin{subfigure}{0.18\textwidth}
        \includegraphics[width=\textwidth]{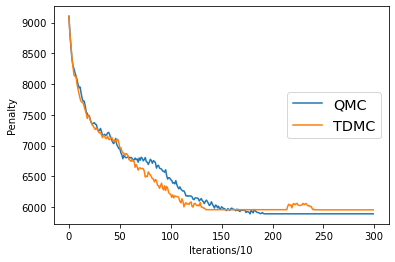}
        \caption{1400 base stations}
        \label{fig:1400basestations}
    \end{subfigure}
    \hfill
    \begin{subfigure}{0.18\textwidth}
        \includegraphics[width=\textwidth]{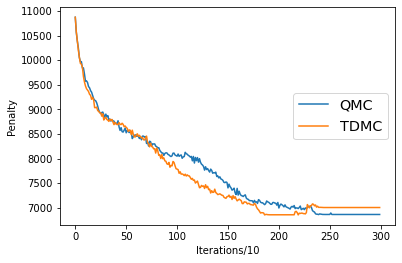}
        \caption{1700 base stations}
        \label{fig:1700basestations}
    \end{subfigure}
    \hfill
    \begin{subfigure}{0.18\textwidth}
        \includegraphics[width=\textwidth]{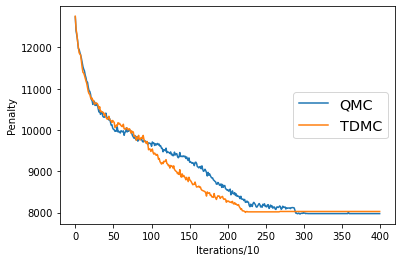}
        \caption{2000 base stations}
        \label{fig:2000basestations}
    \end{subfigure}
    \hfill
    \begin{subfigure}{0.18\textwidth}
        \includegraphics[width=\textwidth]{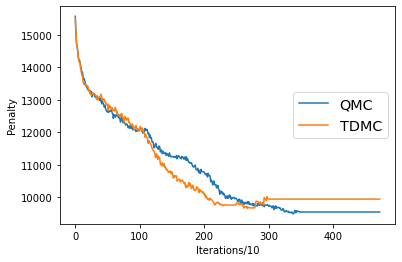}
        \caption{2400 base stations}
        \label{fig:2400basestations}
    \end{subfigure}
    \hfill
    \begin{subfigure}{0.18\textwidth}
        \includegraphics[width=\textwidth]{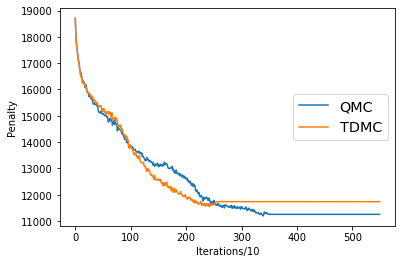}
        \caption{2798 base stations}
        \label{fig:2798basestations}
    \end{subfigure}
    \caption{Implementing QMC and TDMC to find minimum edge servers for different number of base stations}
    \label{fig:Q-learning1stdmodel}
\end{figure*}
\subsection{Implementation Results}
Based on our proposed algorithms, namely QMC and TDMC, the agent could find the near-optimal actions for different numbers of base stations, and the convergence is guaranteed after enough iterations (Fig.~\ref{fig:Q-learning1stdmodel}). The cost, the y-axis, represents the completed path's penalty in each episode. As illustrated, the performances of our two algorithms are relatively similar. However, when the number of base stations becomes big ($> 300$), QMC shows slightly better results and can reach less penalty in comparison with TDMC ($\lambda = 0.4$). Fluctuations in the cost value in the first iterations show the learning process of the agent through trial and error. When the number of base stations increases, the action space expands, and consequently, more iterations are needed to reach the optimal actions.

Although only the fixed part of our state definition is used in creating the Q-table, the state-action values have converged. Interestingly, our algorithms worked for a large number of base stations (Fig.~\ref{fig:1400basestations}-\ref{fig:2798basestations}) with Q-table without the help of neural networks by keeping a part of the state definition as the memory in each episode.

\begin{figure*}[t]
    \centering
    \begin{subfigure}{0.18\textwidth}
        \includegraphics[width=\textwidth]{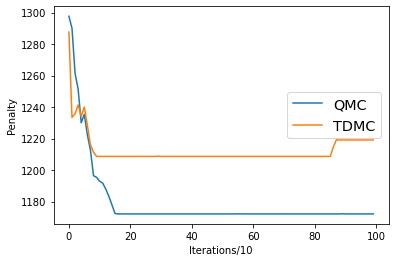}
        \caption{Distance threshold = 3\textit{km}}
        \label{fig:300-3km}
    \end{subfigure}
    \hfill
    \begin{subfigure}{0.18\textwidth}
        \includegraphics[width=\textwidth]{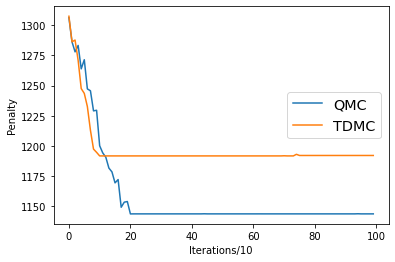}
        \caption{Distance threshold = 5\textit{km}}
        \label{fig:300-5km}
    \end{subfigure}
    \hfill
    \begin{subfigure}{0.18\textwidth}
        \includegraphics[width=\textwidth]{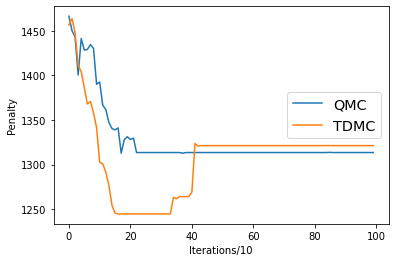}
        \caption{Distance threshold = 7\textit{km}}
        \label{fig:300-7km}
    \end{subfigure}
    \hfill
    \begin{subfigure}{0.18\textwidth}
        \includegraphics[width=\textwidth]{images/300-compQTD1.png}
        \caption{Distance threshold = 9\textit{km}}
        \label{fig:300-9-150km}
    \end{subfigure}
    \hfill
    \begin{subfigure}{0.18\textwidth}
        \includegraphics[width=\textwidth]{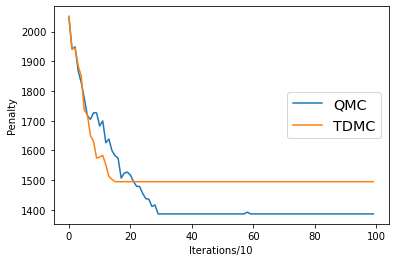}
        \caption{Distance threshold = 11\textit{km}}
        \label{fig:300-11km}
    \end{subfigure}
    \caption{Performance of our proposed algorithm by considering different accepted network delays}
    \label{fig:Q-learning2ndmodeldelayvary}
\end{figure*}

\begin{figure*}
    \centering
    \begin{subfigure}{0.18\textwidth}
        \includegraphics[width=\textwidth]{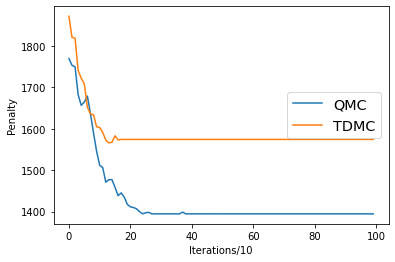}
        \caption{Computation capacity = 100\textit{requests/day}}
        \label{fig:300-9-100}
    \end{subfigure}
    \hfill
    \begin{subfigure}{0.18\textwidth}
        \includegraphics[width=\textwidth]{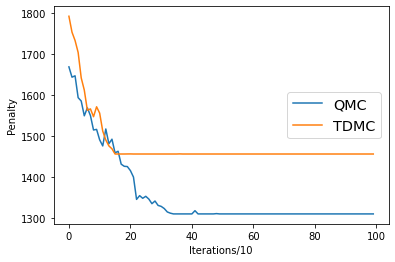}
        \caption{Computation capacity = 120\textit{requests/day}}
        \label{fig:300-9-120}
    \end{subfigure}
    \hfill
    \begin{subfigure}{0.18\textwidth}
        \includegraphics[width=\textwidth]{images/300-compQTD1.png}
        \caption{Computation capacity = 150\textit{requests/day}}
        \label{fig:300-9-150}
    \end{subfigure}
    \hfill
    \begin{subfigure}{0.18\textwidth}
        \includegraphics[width=\textwidth]{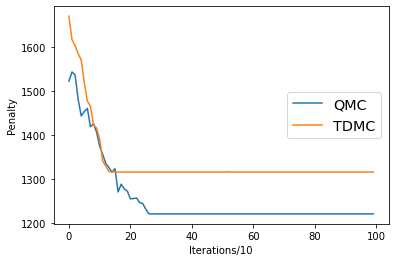}
        \caption{Computation capacity = 170\textit{requests/day}}
        \label{fig:300-9-170}
    \end{subfigure}
    \hfill
    \begin{subfigure}{0.18\textwidth}
        \includegraphics[width=\textwidth]{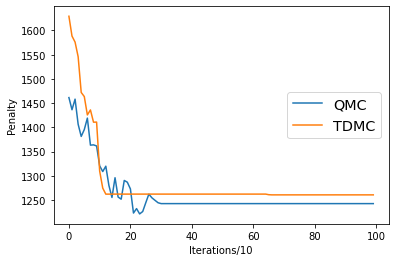}
        \caption{Computation capacity = 200\textit{requests/day}}
        \label{fig:300-9-200}
    \end{subfigure}
    \caption{Performance of our proposed algorithm by considering different computation capacity for the edge servers}
    \label{fig:Q-learning2ndmodelWLvary}
\end{figure*}

Fig.~\ref{fig:Q-learning2ndmodeldelayvary} demonstrates the convergence of our algorithms while the distance threshold varies from 3~\textit{km} to 11~\textit{km}. It means the maximum accepted latency varies between 0.01~\textit{ms} and 0.036~\textit{ms}. This experiment is done for 300 base stations, and the maximum workload is considered 150 requests per day. By increasing the distance threshold, more iterations are needed because the action space is increasing, as well. In the other experiment, the convergence of the model is shown for 300 base stations and the maximum acceptable delay of 0.03~\textit{ms}, while the workload's limit changes between 100 to 200 requests per day (Fig.~\ref{fig:Q-learning2ndmodelWLvary}). Increasing the workload capacity does not necessarily increase the convergence time (comparing Figs.~\ref{fig:300-9-120} and~\ref{fig:300-9-150}), and is dependent on the node's position and the number of neighbors. These two experiments (Figs.~\ref{fig:Q-learning2ndmodeldelayvary}-\ref{fig:Q-learning2ndmodelWLvary}) show that as the distance threshold increases, the performances of two algorithms become more similar.

\subsection{Performance Evaluation}
\subsubsection{TDMC vs QMC with varying constraints}
It is supposed that the processing time is negligible, and by following the workload condition, there is no queue, and consequently, there is no waiting time. Thus, the access delay represents the network delay. Therefore, to evaluate the performance of different algorithms, first, the average distance that represents the communication or access delay and the number of selected edge servers will be repoprted. Then, the combination cost which is the summation of total delay and the cost of adding new edge servers will be analyzed to give us a better view while considering the both objective functions (Eqs.~\ref{eqn:O1}-\ref{eqn:O2}). Before going to the results, a brief explanation of the algorithms is given.

In Top-K, the nodes based on their workloads are sorted, and the first one, with the highest workload, is selected as the edge server. Then the neighbors of the selected edge server are defined. It includes all the nodes that their distance is less than the pre-defined threshold. The neighbor nodes offload their computation to the elected edge server until reaching the maximum capacity of the server. This cluster of the edge server and its connected base stations is removed from the initial set of base stations. Then the remaining nodes are again sorted based on their workloads, and the algorithm is repeated until all the base stations are assigned to an edge server.
Top-DoF and Random algorithms, have the same structure but the edge servers' selecting criteria are different. In Top-DoF, the nodes are sorted based on the number of their neighbors. The node with the most neighbors is selected as the edge server. In the Random algorithm, the first node as the edge server is picked randomly. In both, then the adjacent nodes (neighbors) are mapped to the selected node until reaching the maximum accepted workload. This procedure continues till all nodes are allocated to an edge server.
K-means, where K should be defined in advanced, is one of the most common unsupervised algorithms. As the objective of our solution is to find the number of edge servers, i.e. K, we implemented K-means repetitively by increasing the parameter K until satisfying the load and distance constraints.
The other algorithm is the combination of K-means and Top-K (KmTK). In this method, K-means is executed first until meeting the distance constraint in each cluster. Then, the node having the highest load is selected as the head of each cluster (edge server), before running the Top-K  at the end. Finally, in the genetic algorithm (GA), we implemented the solution that generates random populations and applies mutation and crossover with the same input, cost function, and output that we use in TDMC and QMC.

\begin{table*}
\centering
\caption{The network delay (km) with respect to the number of base stations}
\label{table:comd}
\begin{tabular}{||c | c | c | c | c | c | c | c | c||} 
 \hline
 n & QMC & TDMC & K-means(Km) &Top-K(TK) & KmTK & GA & Top-DoF & Random \\
 \hline\hline
 100 & 2.74 & 2.46 & $\boldsymbol{0.1}$ & 3.8 & 1.97 & 2.87 & 4.44 & 3.47 \\ 
 \hline
 300 & 2.48 & 2.42 & $\boldsymbol{0.1}$ & 4 & 2.89 & 3.22 & 4.64 & 4.26 \\
 \hline
 500 & 2.33 & 2.48 & $\boldsymbol{0}$ & 4.11 & 2.47 & 3.33 & 4.27 & 4.15 \\
 \hline
 700 & 2.26 & 2.41 & $\boldsymbol{0}$ & 4.16 & 2.34 & 3.53 & 4.55 & 4.29 \\
 \hline
 1000 & 2.21 & 2.2 & $\boldsymbol{0}$ & 4.32 & 2.38 & 3.4 & 4.49 & 4.36 \\
 \hline
 1400 & 2.13 & 2.19 & $\boldsymbol{0}$ & 4.4 & 2.02 & 3.82 & 4.47 & 4.36 \\
 \hline
 1800 & 2.14 & 2.23 & $\boldsymbol{0}$ & 4.42 & 1.96 & 3.9 & 4.52 & 4.5 \\
 \hline
 2000 & 2.16 & 2.2 & $\boldsymbol{0}$ & 4.55 & 2.04 & 3.96 & 4.49 & 4.66\\
 \hline
 2400 & 2.18 & 2.31 & $\boldsymbol{0}$ & 5.09 & 1.91 &  4.14 & 4.56 & 5.02\\
 \hline
 2798 & 2.19 & 1.36 & $\boldsymbol{0}$ & 5.59 & 1.11 &  4.17 & 4.58 & 5.42\\
 \hline
\end{tabular}
\end{table*}
\begin{table*}[t]
\centering
\caption{The number of selected edge servers with respect to the number of base stations}
\label{table:comn}
\begin{tabular}{||c | c | c | c | c | c | c | c | c||} 
 \hline
 n & QMC & TDMC & K-means(Km) &Top-K(TK) & KmTK & GA & Top-DoF & Random \\
 \hline\hline
 100 & $\boldsymbol{17}$ & 18 & 69 & 19 & 29 & 18 & 18 & 18 \\ 
 \hline
 300 & 59 & 59 & 208 & 63 & 93 & $\boldsymbol{57}$ & 63 & 64 \\
 \hline
 500 & 109 & 109 & 469 & 114 & 226 & $\boldsymbol{107}$ & 112 & 114 \\
 \hline
 700 & 171 & 171 & 672 & 176 & 342 & $\boldsymbol{170}$ & 172 & 175 \\
 \hline
 1000 & $\boldsymbol{224}$ & 225 & 968 & 230 & 470 & $\boldsymbol{224}$ & 228 & 230 \\
 \hline
 1400 & 285 & 284 & 1346 & 290 & 621 & $\boldsymbol{278}$ & 288 & 292 \\
 \hline
 1800 & 318 & 318 & 1697 & 323 & 706 &  $\boldsymbol{312}$ & 324 & 328 \\
 \hline
 2000 & 360 & 358 & 1895 & 363 & 851 &  $\boldsymbol{352}$ & 364 & 365\\
 \hline
 2400 & 428 & 428 & 2282 & 433 & 1000 &  $\boldsymbol{422}$ & 435 & 443\\
 \hline
 2798 & 509 & 510 & 2662 & 512 & 1216 &  $\boldsymbol{501}$ & 517 & 521\\
 \hline
\end{tabular}
\end{table*}
\begin{figure*}
    \centering
    \includegraphics[width=\textwidth, height = 8cm]{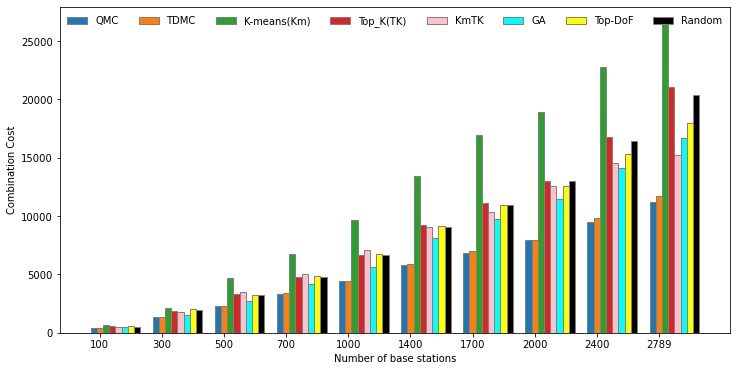}
    \caption{Cost comparison by considering the two objectives for different number of base stations}
    \label{fig:Combcost}
\end{figure*}
\subsubsection{{Network delay, edge servers, and cost with varying number of base stations}}
Tables ~\ref{table:comd}-\ref{table:comn} illustrate the performance of the considered algorithms for different number of base stations ranging from 100 to 2798. Each table represents one of the objectives. The network delay for repetitive K-means is zero in most cases. On the other hand, the number of edge servers is close to the number of base stations, which is similar to the situation when each base station is selected as the edge server. The combination of K-means and Top-K (KmTK) has reduced the network delay compared to the outcomes of Top-K. However, the number of edge servers has increased. KmTK has less delay compared to QMC and TDMC, but it has more number of edge servers with a significant difference. The performance of the Random is interesting, and in most cases, it has some striking similarities to the performance of the Top-K in terms of network delay. Considering the number of edge servers, GA has the minimum cost while the network delay is higher than some other methods. Therefore, by comparing these two tables, creating a combination cost is crucial to have a holistic view. 


\begin{equation}\label{eqn:combcost}
    combination\_cost = \sum_{i=1}^{n} \sum_{j=1}^{K} d_{ij} + 10 \times K
\end{equation}
In Eq.~\ref{eqn:combcost}, $K$ is the number of the edge servers and as the maximum accepted delay is 9km, the cost of adding each edge server is considered equal to 10. 
Fig.~\ref{fig:Combcost} shows the combination cost of all the algorithms for different number of base stations. The combination cost of QMC and TDMC is less than the other algorithms, while for 100 and 300 base stations, TDMC performs batter. In the other cases, TDMC and QMC are very close. After these two algorithms, GA shows in most of the cases better performance compared to the rest of the algorithms because both objectives are considered simultaneously. Up to 2000 base stations, KmTK, Top-K, Top-DoF and Random reveal very close outcomes. When the number of base stations increases, KmTK shows better cost followed by Top-Dof. Top-K and Random show similar trends. Finally, K-means is ranked last.

\subsubsection{Network delay, edge servers, and cost with varying distance}
In the other experiment, the number of base stations is 300, and the maximum computation capacity of edge servers is 150 requests per day (Tables~\ref{table:comdd}-\ref{table:comnd}). These two factors are fixed, and the distance constraint varies. The performance of algorithms is compared in terms of the number of edge servers ($K$) in Table~\ref{table:comnd}, and in terms of network delay $(D_{N})$ in Table~\ref{table:comdd}, which represents the quality of the network. 

%
\begin{table*}[t]
\centering
\caption{The network delay (km) with respect to the distance constraint for 300 base stations}
\label{table:comdd}
\begin{tabular}{||c | c | c | c | c | c | c | c | c||} 
 \hline
 $D_{TH}$ & QMC & TDMC & K-means(Km) &Top-K(TK) & KmTK & GA & Top-DoF & Random \\
 \hline\hline
 3 & 1.1 & 1.07 & $\boldsymbol{0.076}$ & 1.21 & 0.71 & 1.25 & 1.31 & 1.18 \\ 
 \hline
 5 & 1.48 & 1.6 & $\boldsymbol{0.077}$ & 1.98 & 1.74 & 1.88 & 2.17 & 2.14 \\
 \hline
 7 & 2.18 & 2.22 & $\boldsymbol{0.093}$ & 2.94 & 1.92 & 2.59 & 3.34 & 3.01 \\
 \hline
 9 & 2.48 & 2.42 & $\boldsymbol{0.108}$ & 4 & 2.89 & 3.22 & 4.64 & 4.26 \\
 \hline
 11 & 2.65 & 3 & $\boldsymbol{0.112}$ & 4.87 & 3.01 & 3.8 & 5.55 & 4.65\\
 \hline
\end{tabular}
\end{table*}
\begin{table*}[t]
\centering
\caption{The number of selected edge servers with respect to the distance constraint for 300 base stations}
\label{table:comnd}
\begin{tabular}{||c | c | c | c | c | c | c | c | c||} 
 \hline
 $D_{TH}$ & QMC & TDMC & K-means(Km) &Top-K(TK) & KmTK & GA & Top-DoF & Random \\
 \hline\hline
 3 & $\boldsymbol{79}$ & 82 & 224 & 94 & 133 & 80 & 88 & 91 \\ 
 \hline
 5 & 67 & 67 & 220 & 74 & 108 & $\boldsymbol{63}$ & 74 & 72 \\
 \hline
 7 & 62 & 61 & 213 & 67 & 102 & $\boldsymbol{60}$ & 69 & 64 \\
 \hline
 9 & 59 & 59 & 208 & 63 & 93 & $\boldsymbol{57}$ & 63 & 64 \\
 \hline
 11 & 57 & 57 & 206 & 61 & 91 &  $\boldsymbol{56}$ & 61 & 62\\
 \hline
\end{tabular}
\end{table*}
\begin{figure*}
    \centering
    \includegraphics[width=\textwidth, height = 8cm]{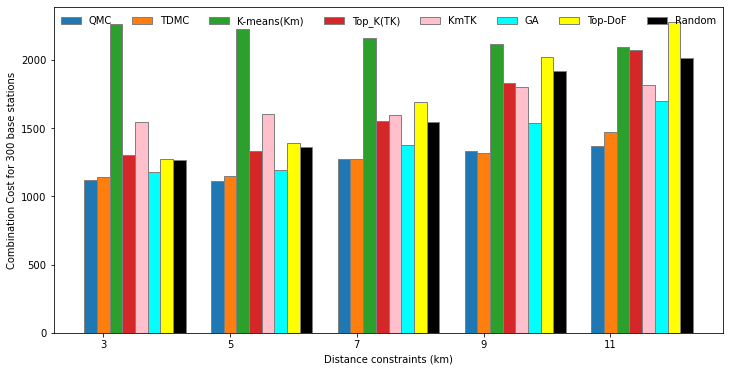}
    \caption{Cost comparison by considering the two objectives for different distance constraints}
    \label{fig:Combcostexp2}
\end{figure*}
Less delay, as an integral feature of the MEC, results in higher QoS and customer satisfaction. It is clear that increasing the distance constraint results in higher latency for all the algorithms. In most of the cases, after K-means, our QMC and TDMC algorithms have less delay (Table~\ref{table:comdd}).

The number of selected edge servers decreases as the delay increases, except in one case for the Random algorithm. The Random algorithm has exceptional results compared to Top-K and Top-DoF, and it outperforms in some situations. Similar to the previous experience, GA has reached the minimum number of edge servers except for 3$km$ distance threshold, where QMC has fewer edge servers (Table~\ref{table:comnd}).

As shown in Fig.~\ref{fig:Combcostexp2}, QMC and TDMC have the lowest combination cost for different distances. Then GA followed by Top-K show better cost compared to the rest of algorithms for small distances. As the distance constraint becomes less tight, KmTK outperforms Top-K, but GA still reveals better cost. We can also observe that the cost of K-means decreases while increasing the distance to the point that this cost becomes close to the one generated by Top-K and less than the cost of Top-DoF when $D_{th}=11 km$. In fact, although Top-DoF shows competitive cost for small distances, the approach becomes expensive for long distances.

\subsubsection{Network delay, edge server, and cost with varying workload}
In our last experiment, the effect of workload constraint is investigated. Similar to the previous one, 300 base stations are considered when the maximum accepted delay is 0.03~\textit{ms}, but here different edge servers capacities are examined.
\begin{table*}[t]
\centering
\caption{The network delay (km) with respect to the edge servers computation capacity for 300 base stations}
\label{table:comdl}
\begin{tabular}{||c | c | c | c | c | c | c | c | c||} 
 \hline
 $\Delta_{TH}$ & QMC & TDMC & K-means(Km) &Top-K(TK) & KmTK & GA & Top-DoF & Random \\
 \hline\hline
 100 & 1.46 & 2.17 & $\boldsymbol{0.01}$ & 3.2 & 1.97 & 3.13 & 3.55 & 3.22 \\ 
 \hline
 120 & 1.87 & 2.28 & $\boldsymbol{0.07}$ & 3.37 & 2.46 & 3.24 & 3.77 & 3.12 \\
 \hline
 150 & 2.48 & 2.42 & $\boldsymbol{0.108}$ & 3.39 & 2.89 & 3.22 & 4.02 & 3.16 \\
 \hline
 170 & 2.17 & 2.57 & $\boldsymbol{0.18}$ & 3.36 & 2.93 & 3.36 & 3.91 & 3.55 \\
 \hline
 200 & 2.47 & 2.74 & $\boldsymbol{0.29}$ & 3.56 & 2.84 & 3.43 & 4.2 & 3.42\\
 \hline
\end{tabular}
\end{table*}
\begin{table*}[t]
\centering
\caption{The number of selected edge servers with respect to the edge servers computation capacity for 300 base stations}
\label{table:comnl}
\begin{tabular}{||c | c | c | c | c | c | c | c | c||} 
 \hline
 $\Delta_{TH}$ & QMC & TDMC & K-means(Km) &Top-K(TK) & KmTK & GA & Top-DoF & Random \\
 \hline\hline
 100 & 93 & 85 & 288 & 91 & 131 & $\boldsymbol{82}$ & 91 & 92 \\ 
 \hline
 120 & 75 & 71 & 249 & 75 & 123 & $\boldsymbol{69}$ & 76 & 79 \\
 \hline
 150 & 59 & 59 & 208 & 63 & 93 & $\boldsymbol{57}$ & 63 & 64 \\
 \hline
 170 & 54 & 52 & 189 & 58 & 79 & $\boldsymbol{50}$ & 54 & 64 \\
 \hline
 200 & 47 & 42 & 166 & 48 & 65 &  $\boldsymbol{42}$ & 47 & 61\\
 \hline
\end{tabular}
\end{table*}
\begin{figure*}
    \centering
    \includegraphics[width=\textwidth, height = 8cm]{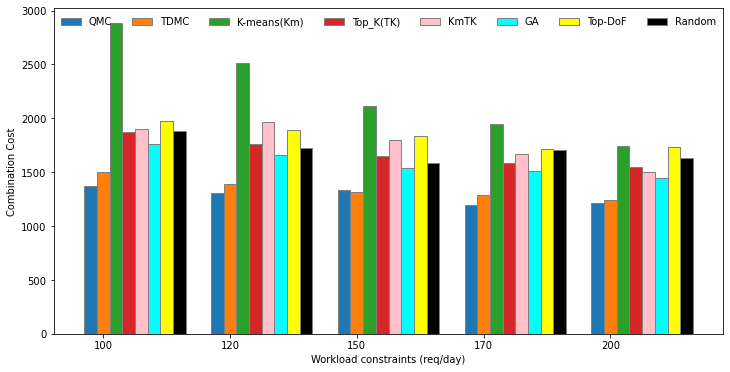}
    \caption{Cost comparison by considering the two objectives for different edge servers computation capacity}
    \label{fig:Combcostexp3}
\end{figure*}

Concerning the network delay, after K-means, QMC noticeably outperforms the others benchmark approaches (Table~\ref{table:comdl}). The TDMC and then KmTK algorithms, on average, have a better performance compared to Random, Top-K and Top-DoF. To find the minimum number of edge servers, as the capacity increases, the number of edge servers decreases. GA has the best performance for finding the smallest number of edge servers, then TDMC outperforms the other algorithms
(Table~\ref{table:comnl}). As the tables show, for each objective, different algorithms have better performance compared to the others, so we need to compare the combination cost of these algorithms. Fig.~\ref{fig:Combcostexp3} reveals that in all the cases, QMC and TDMC outperform the other solutions, with a slightly better advantage for QMC. The third best algorithm is GA followed by Random and Top-K that, on average, have better performance compare to KmTK and Top-K, while in all cases the K-means has the highest cost.

Based on all these experiments, it can be concluded that QMC, TDMC and GA have less cost because these algorithms consider both objectives: network delay and number of edge servers  simultaneously.

\subsection{The Selection of Parameters $\alpha$ and $\gamma$}
In our experiments, deciding on the values of $\alpha$ and $\gamma$ is critical for the agent performance in terms of quality of the decisions. In Table \ref{tab:parameters_tuning}, we show 
the minimum cost achieved by the agent while varying $\alpha$ and $\gamma$. To this end, we consider $700$ base stations and set the maximum distance to be $5$ km with $120$ requests per day as a workload.
\begin{table}[]
    \centering
    \begin{tabular}{|c|c|c|}
    \hline
        $\alpha$ & $\gamma$ & Cost \\
        \hline
        \hline
        0.9 & 0.5 & 2982.8254\\
        \hline
        $\boldsymbol{0.4}$ & $\boldsymbol{0.5}$ & $\boldsymbol{2982.8254}$\\
        \hline
        0.4 & 0.9 & 2967.9766\\
        \hline
        0.2 & 0.9 & 2972.5821\\
        \hline
        0.2 & 0.2 & 2982.8254\\
        \hline
        0.7 & 0.7 & 2982.8254\\
        \hline
        0.7 & 0.9 & 2982.8254\\
        \hline
        $\boldsymbol{0.8}$& $\boldsymbol{0.3}$ &$\boldsymbol{2967.9766}$\\
        \hline
    \end{tabular}
    \caption{The impact of $\alpha$ and $\gamma$ on the performance of the agent. $n = 700$; $D_{TH} = 5$ km; $120$ request/day}
    \label{tab:parameters_tuning}
\end{table}

Based on the results shown in Table \ref{tab:parameters_tuning}, with higher values of $\alpha$, the agent takes bigger steps in the learning process with a probability of missing some better actions, which could result in lower costs. When $\alpha$ is too small, it takes more iterations to converge and the results also might not be satisfying. The magnitude of $\gamma$ shows the importance of future rewards. Both of these hyperparameters should be defined based on each problem/input size. In addition, we can see that the minimum cost is obtained in two settings. Therefore, our decision for choosing $\alpha = 0.4$ and $\gamma = 0.9$ is based on trial and error for different values of these two parameters.

\subsection{Deep Reinforcement Learning} \label{DRL-DQN}
\begin{figure}
    \centering
    \includegraphics[scale=0.6]{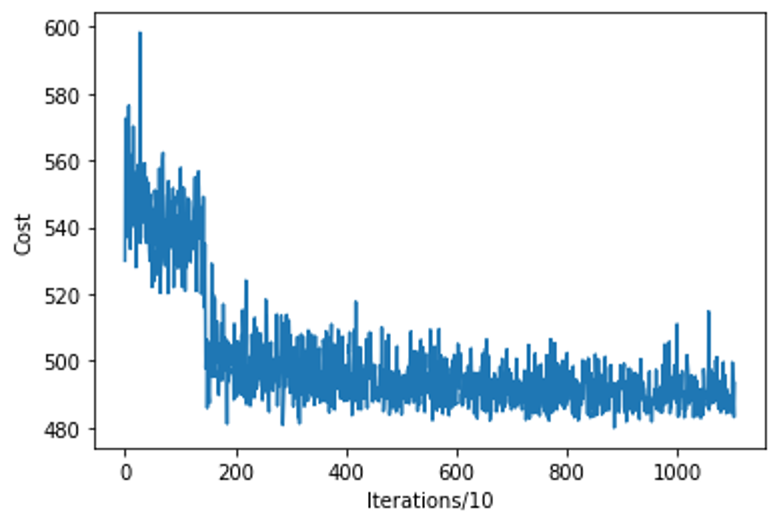}
    \caption{DQN performance with 100 base stations}
    \label{fig:exp_drl}
\end{figure}

In this experiment, we evaluate the performance of  DQN to solve our edge server placement problem, while considering the same constraints and MDP design. The performance of the DQN agent is shown in Fig. \ref{fig:exp_drl}.
In this figure, we can notice that DQN is not able to converge to a smaller cost or stabilize the decisions compared to QMC and TDMC (see Fig. \ref{fig:100basestations}). The main reason behind the poor performance of DQN is that our efficient MDP design does not require a non-linear approximation that DQN performs to approximate the Q-table. Implementing this non-necessary approximation causes the model to generate and select non-optimal actions, which results in unstable cost values. Moreover, DQN requires a high number of parameters that require further training and careful tuning.

\section{Conclusions}\label{section5}

In this paper, a new RL-based algorithm is developed 
for jointly optimizing the placement and computation offloading to minimize the cost of the MEC design. The challenges of conducting this research are the limitations of RL when used for real-world applications, including the consideration of variable action space, difficulty in penalty definition for multi-objective problems, lack of a specific static end goal, and the curse of dimensionality caused by massive action space and state space explosion. Our RL-based frameworks provide an efficient solution for this joint problem while considering these issues. Despite the mentioned challenges, the key to Q-learning application for this large-scale real-world problem is a precise and efficient representation of the state space, action space, and penalty function. In addition of obtaining a convergent solution for our proposed algorithms, which is a difficult challenge in RL, we have proposed an efficient formalization that resulted in promising outcomes. The performance obtained from our algorithms demonstrated that the RL-based solutions have better performance in cost reduction, which includes both reducing network delay and the number of the edge servers, compared to relevant benchmark methods. We also showed that applying a deep reinforcement solution resulted in a poor outcome since the MDP design does not require a non-linear approximation. For future work, we are investigating the ability to generalize the algorithm for larger and unseen networks by defining the MEC as a graph and then deploying a graph representation  to encode the state and action spaces.

\ifCLASSOPTIONcaptionsoff
  \newpage
\fi



%

\bibliography{mybib}
\bibliographystyle{IEEEtran}
\begin{IEEEbiography}
[{\includegraphics[width=1in,height=1.25in,clip,keepaspectratio]{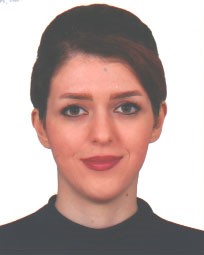}}]{Anahita Mazloomi}
received her second M.Sc. from Concordia University, Institute for information Systems Engineering (CIISE). She received her first M.Sc. degree in industrial engineering from the Kharazmi University of Tehran and completed her B.S. at the Yazd University. The topics of her research are Edge Computing and Reinforcement Learning.
\end{IEEEbiography}
\vskip 0pt plus -1fil
\begin{IEEEbiography}
[{\includegraphics[width=1in,height=1.25in,clip,keepaspectratio]{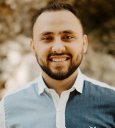}}]{Hani Sami}
is currently pursuing his Ph.D. at Concordia University, Institute for information Systems Engineering (CIISE). He received his M.Sc. degree in Computer Science from the American University of Beirut  and completed his B.S. and worked as Research Assistant at the Lebanese American University. The topics of his research are fog computing, vehicular fog computing, and reinforcement learning. He is a reviewer of several prestigious conferences and journals.
\end{IEEEbiography}
\vskip 0pt plus -1fil
\begin{IEEEbiography}
[{\includegraphics[width=1in,height=1.25in,clip,keepaspectratio]{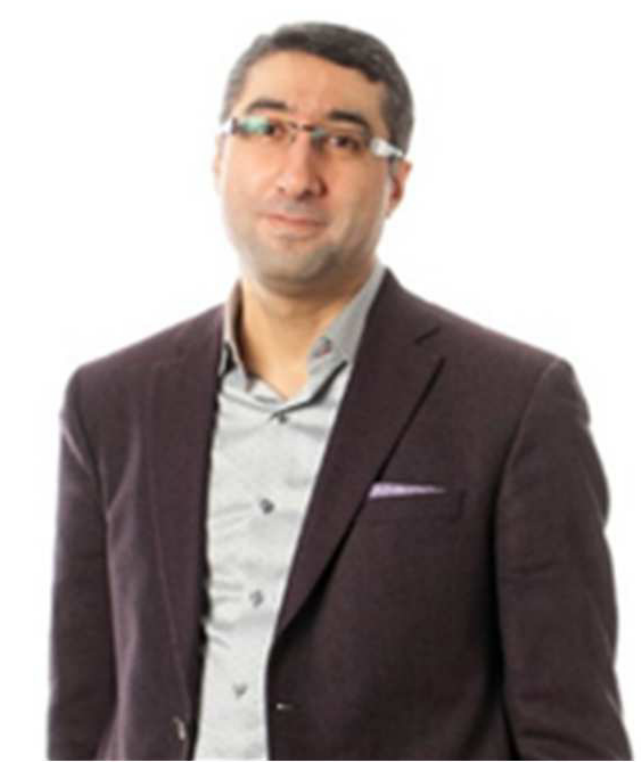}}]{Jamal Bentahar}
received the Ph.D. degree in computer science and software engineering from Laval University, Canada, in 2005. He is a Professor with Concordia Institute for Information Systems Engineering, Concordia University, Canada. From 2005 to 2006, he was a Postdoctoral Fellow with Laval University, and then NSERC Postdoctoral Fellow at Simon Fraser University, Canada. He is an NSERC Co-Chair for Discovery Grant for Computer Science (2016-2018). His research interests include the areas of computational logics, reinforcement learning, multi-agent systems, service computing, game theory, and software engineering.
\end{IEEEbiography}
\vskip 0pt plus -1fil
\begin{IEEEbiography}
[{\includegraphics[width=1in,height=1.25in,clip,keepaspectratio]{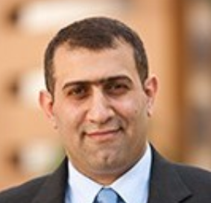}}]{Hadi Otrok}
holds an associate professor position in the department of ECE at Khalifa University of Science and Technology, an affiliate associate professor in the Concordia Institute for Information Systems Engineering at Concordia University, Montreal, Canada, and an affiliate associate professor in the electrical department at Ecole de Technologie Superieure (ETS), Montreal, Canada. He received his Ph.D. in ECE from Concordia University. He is a senior member at IEEE, and associate editor at: Ad-hoc networks (Elsevier) and IEEE Networks. He served in the editorial board of IEEE communication letters. He co-chaired several committees at various IEEE conferences. His research interests include the domain of computer and network security, crowd sensing and sourcing, ad hoc networks, reinforcement learning, and blockchain.
\end{IEEEbiography}
\vskip 0pt plus -1fil
\begin{IEEEbiography}
[{\includegraphics[width=1in,height=1.25in,clip,keepaspectratio]{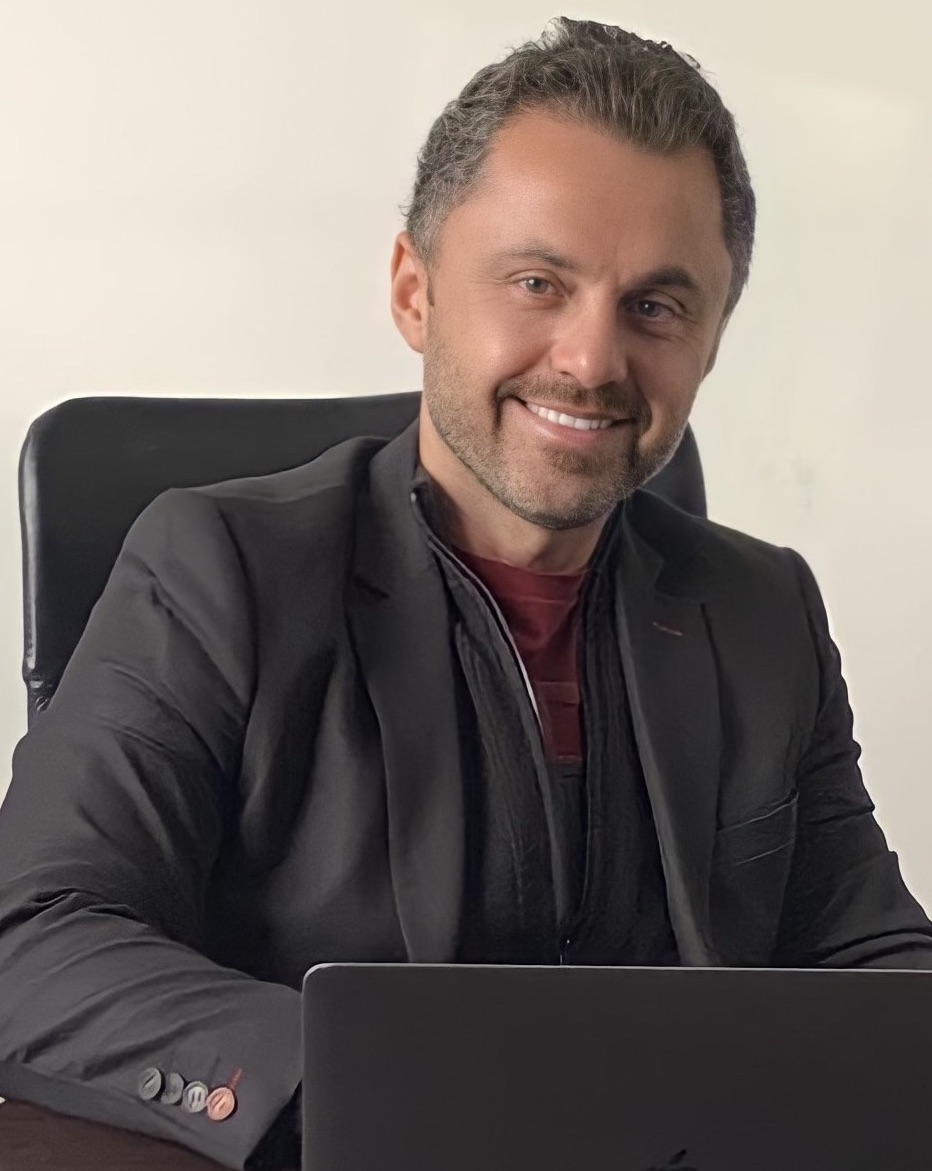}}]{Azzam Mourad}
received his M.Sc. in CS from Laval University, Canada (2003) and Ph.D. in ECE from Concordia University, Canada (2008). He is currently an associate professor of computer science with the Lebanese American University and an affiliate associate professor with the Software Engineering and IT Department, Ecole de Technologie Superieure (ETS), Montreal, Canada. He published more than 100 papers in international journal and conferences on Security, Network and Service Optimization and Management targeting IoT, Cloud/Fog/Edge Computing, Vehicular and Mobile Networks, and Federated Learning. He has served/serves as an associate editor for IEEE Transaction on Network and Service Management, IEEE Network, IEEE Open Journal of the Communications Society, IET Quantum Communication, and IEEE Communications Letters, the General Chair of IWCMC2020, the General Co-Chair of WiMob2016, and the Track Chair, a TPC member, and a reviewer for several prestigious journals and conferences. He is an IEEE senior member.
\end{IEEEbiography}

\end{document}